\documentclass{article}
\usepackage{amsmath,graphicx,mlspconf}
\usepackage{xcolor}
\usepackage{tabularx}
\usepackage{booktabs}
\usepackage{amsmath}
\usepackage{amsfonts}
\usepackage{url}
\usepackage{multirow}

\toappear{2025 IEEE International Workshop on Machine Learning for Signal Processing, Aug.\ 31-- Sep.\ 3, 2025, Istanbul, Turkey}

\title{Uncertainty Quantification for Motor Imagery BCI --\\ Machine Learning vs. Deep Learning}
\name{Anonymous\thanks{Anonymous.}}
\address{Anonymous}

\name{Joris Suurmeijer$^{*}$\thanks{$^{*}$ These authors contributed equally to this work.} \qquad Ivo Pascal de Jong$^{*}$ \qquad Matias Valdenegro-Toro \qquad Andreea Ioana Sburlea}

\address{%
   Bernoulli Institute, University of Groningen, The Netherlands \\
   Email: ivo.de.jong@rug.nl
}

\begin{document}

\maketitle

\begin{abstract}
Brain-computer interfaces (BCIs) turn brain signals into functionally useful output, but they are not always accurate.  A good Machine Learning classifier should be able to indicate how confident it is about a given classification, by giving a probability for its classification. 
Standard classifiers for Motor Imagery BCIs do give such probabilities, but research on uncertainty quantification has been limited to Deep Learning. We compare the uncertainty quantification ability of established BCI classifiers using Common Spatial Patterns (CSP-LDA) and Riemannian Geometry (MDRM) to specialized methods in Deep Learning (Deep Ensembles and Direct Uncertainty Quantification) as well as standard Convolutional Neural Networks (CNNs). 

We found that the overconfidence typically seen in Deep Learning is not a problem in CSP-LDA and MDRM. We found that MDRM is underconfident, which we solved by adding Temperature Scaling (MDRM-T). CSP-LDA and MDRM-T give the best uncertainty estimates, but Deep Ensembles and standard CNNs give the best classifications. We show that all models are able to separate between easy and difficult estimates, so that we can increase the accuracy of a Motor Imagery BCI by rejecting samples that are ambiguous.

\end{abstract}

\newcommand{\cem}[1]{\textcolor{blue}{cem: #1}}

\section{Introduction} 	\label{sec:introduction}

Research on non-invasive BCIs relies on strong Machine Learning algorithms that can accurately detect cognitive states from EEG. In this direction, a lot of work is done to improve the classification accuracy of models under various circumstances. However, no model will ever be perfect, which makes it desirable to know how \textit{confident} the model is about a given classification, so that we can know whether to trust that classification. So far, the literature on EEG-based Motor Imagery BCIs has only given this aspect very little attention \cite{manivannan2024uncertainty, milanes2023robust}.

\begin{figure}[t]
    \begin{minipage}[t]{0.31\linewidth}
      \centering
      \centerline{\includegraphics[width=\linewidth]{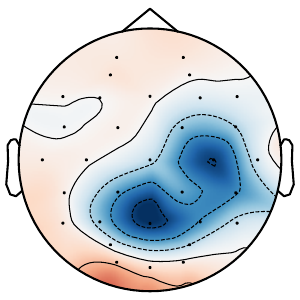}}
      \centerline{\texttt{LEFT - \textbf{95\%}}}
    \end{minipage}\hfill
    \begin{minipage}[t]{0.31\linewidth}
      \centering
      \centerline{\includegraphics[width=\linewidth]{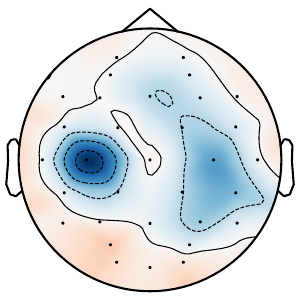}}
      \centerline{\texttt{RIGHT - \textbf{98\%}}}
    \end{minipage}\hfill
    \begin{minipage}[t]{0.31\linewidth}
      \centering
      \centerline{\includegraphics[width=\linewidth]{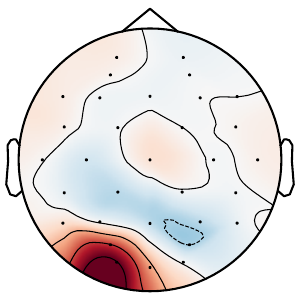}}
      \centerline{\texttt{UNCERTAIN \textbf{<65\%}}}
    \end{minipage}

    \begin{minipage}[t]{0.31\linewidth}
      \centering
      \centerline{\includegraphics[width=\linewidth]{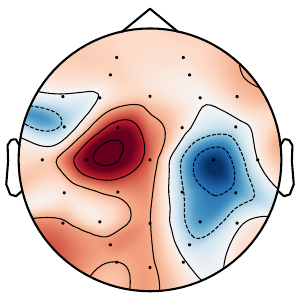}}
      \centerline{\texttt{LEFT - \textbf{72\%}}}
    \end{minipage}\hfill
    \begin{minipage}[t]{0.31\linewidth}
      \centering
      \centerline{\includegraphics[width=\linewidth]{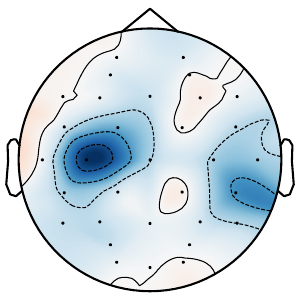}}
      \centerline{\texttt{RIGHT - \textbf{90\%}}}
    \end{minipage}\hfill
    \begin{minipage}[t]{0.31\linewidth}
      \centering
      \centerline{\includegraphics[width=\linewidth]{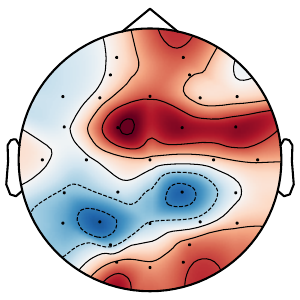}}
      \centerline{\texttt{UNCERTAIN \textbf{<65\%}}}
    \end{minipage}

    \caption{Examples of Event Related Desynchronisation (blue) and Synchronization (red) trials in a left vs. right hand motor imagery task, with illustrative classifications and probabilities (i.e. confidence) from a model below each. Only when there is a unilateral desynchronization over the motor cortex the model is confident and gives a high probability for the correct class, otherwise it is uncertain. Class probabilities should reflect the probability of being correct, and should be useful for separating between correct and incorrect classifications.}
    \label{fig:uq_goals}
\end{figure}

The field of Uncertainty Quantification (UQ) focuses on the \textit{confidence} of ML models when making predictions. Confidence is defined as the predicted probability of a given prediction from an ML to be correct (where "uncertainty" is simply the opposite of "confidence"). In practice this is the probability you see before applying a threshold to select a class. 
The goal of UQ is to ensure that the confidence of an ML model aligns with the actual accuracy of the model. If the model has a confidence of 70\%, then it should be correct in 70\% of the time. To assess this, the \emph{calibration error} can be measured, which compares the confidence of the ML model to its actual accuracy. Good uncertainty estimation should also allow us to intervene when a given prediction has a low probability of being correct \cite{milanes2023robust}. These desired behaviors are illustrated in Figure \ref{fig:uq_goals}.

In the context of Motor Imagery BCIs, UQ can prevent a device from performing an action when the model used is not confident about the prediction. Otherwise a person using a BCI could send unintended controls to a device. UQ can prevent these unwanted predictions when we only execute controls when the model is, e.g. 80\% confident that it is correct. This requires a strong alignment between accuracy and confidence. This alignment is what we will measure. 

Previous studies on uncertainty mainly consider Deep Learning models \cite{de2023uncertainty}, with an emphasis on Bayesian Neural Networks \cite{milanes2023robust}. While Neural Networks have been shown to get good classification accuracy, they are also known to be overconfident and computationally expensive. The computational cost may be an obstacle for online implementations. Therefore, we investigate the quality of uncertainty estimates from standard Motor Imagery BCI models, specifically looking at Common Spatial Patterns with Linear Discriminant Analysis (CSP-LDA; \cite{blankertz2007optimizing}) and Minimum Distance to Riemannian Mean (MDRM; \cite{barachant2011multiclass}), and compare them to Deep Learning based uncertainty estimates from Deterministic Uncertainty Quantification (DUQ; \cite{van2020uncertainty}) and Deep Ensembles \cite{schirrmeister2017deep}.

Overall this work contributes the first complete comparison between the quality of uncertainty estimates from Deep Learning and traditional Motor Imagery BCI models.

\section{Methods}\label{sec:methods}

In our experiments we use multiple existing datasets from a standard Motor Imagery BCI set-up where participants imagine a cued movement which is then decoded from EEG signals. Specifically, we use the datasets from Steyrl et al. \cite{steyrl2016random}, Zhou et al. \cite{zhou2016fully}, the BCI Competition IV - dataset 2b \cite{leeb2007brain} and BCI Competition IV - dataset 2a \cite{tangermann2012review}.  All datasets were accessed through Mother Of All BCI Benchmarks (MOABB; \cite{moabb}). %
We apply an 80-20 within-subject train-test split. The Deep Learning models use 10\% of the train split as validation data for early stopping. We aim to keep things consistent between methods and datasets by maintaining the same preprocessing pipelines. All data is preprocessed using a single non-causal IIR Band-pass filter, with a frequency band of $7.5-30$Hz. The full code implementation can be found at \url{github.com/Jorissuurmeijer/UQ-motor-imagery}.

\subsection{Uncertainty Quantification Methods}\label{sec:_Models}

We will briefly describe the core concepts underlying the Machine Learning and Uncertainty Quantification methods that are used in this study. For a more comprehensive explanation behind these methods we refer to the original publications.

Minimum Distance to Riemannian Mean (MDRM; \cite{barachant2011multiclass}) is a distance-based classifier that relies on the geometry of covariance matrices, and is well-established in BCI literature. For MDRM each epoch is represented as the covariance matrix between EEG channels, and we learn a mean covariance matrix for each class. The covariance matrices fall on a manifold, and we use Riemannian Geometry to measure distance over this manifold, instead of through euclidean space. Inference is done by finding the Riemannian distances to each class mean, where the closest class is the prediction. To get class probabilities (i.e. confidence) the softmax function is used. 

We propose to improve the uncertainty estimation of MDRM by adding Temperature Scaling \cite{guo2017calibration}. We call this model MDRM-T.  With MDRM-T the distances $d$ are scaled by a temperature parameter $T$, so that class probabilities $P(\hat{y})$ are calculated as 
\begin{equation}
    \hat{y} = \frac{\exp(-d_i^2 / T)}{\sum_{j} \exp(-d_j^2 / T)}.
    \label{eq:softmaxtemp}
\end{equation}
By scaling down all the logits with a large temperature $T$ all the predicted class probabilities come closer together. This means the model's predictions will be more uncertain. With a small temperature $T$ the probabilities are pushed more towards the extremes (0 and 1) and thus the model's predictions will be more confident.
The temperature parameter $T$ is optimized after training the model by finding the value that minimizes the calibration error on the training data. This way the model should be neither overconfident nor underconfident.

Common Spatial Patterns with Linear Discriminant Analysis (CSP-LDA; \cite{blankertz2007optimizing}) is a combination of two Machine Learning methods commonly used in EEG classification. CSP learns linear combinations of channels (filters) to form surrogate sensors. Each filter (here: 8) maximizes the variance for one class while minimizing the variance for the other class(es). We then take the average band power so that each filter becomes a scalar feature for an LDA classifier. LDA finds an optimal axis that samples can then be projected onto, to achieve class separation. Class probabilities are calculated using Bayes' rule by assuming each class follows a multivariate Gaussian distribution.

The Deep Learning based methods rely on the well-established ShallowConvNet architecture \cite{Shallowconvnet}. This is a fairly small Convolutional Neural Network (CNN) which is designed for EEG classification and has shown good results for BCI tasks. We train the ShallowConvNet models with early stopping and the Adam optimizer.%
The Softmax output represents the class probabilities as a measure of confidence. 

Deep Ensembles \cite{lakshminarayanan2017simple} uses five ShallowConvNet models trained on the same data. Through differences in random initialization these models will make slightly different predictions, which increases the uncertainty. This allows the model to represent \textit{model uncertainty}, which is what arises when a model needs to make inferences on data that is dissimilar to what it was trained on. Deep Ensembles are often shown to be the best-performing Uncertainty Quantification method in Deep Learning \cite{de2023uncertainty}. 

Direct Uncertainty Quantification (DUQ; \cite{van2020uncertainty}) uses a distance-based method for estimating uncertainty. It replaces the usual softmax layer with a Radial Basis Function (RBF) kernel (length scale 0.2). It learns a centroid in the output space to represent each class. Inference is based on the distance to each centroid. In the original DUQ implementation the distance to the nearest class is considered the uncertainty, but this does not allow us to observe calibration error. To turn the distance into class probabilities we again use softmax with temperature scaling as shown in Equation \ref{eq:softmaxtemp}, similar to MDRM-T.

\subsection{Metrics} \label{sec:evaluation methods}

For the evaluation of UQ performance, specific metrics are needed that indicate how well-calibrated the ML models are. 

In this study, the Expected Calibration Error (ECE; \cite{guo2017calibration}),  Net Calibration Error (NCE; \cite{groot2024confidence}), and the Brier Score \cite{graf1999assessment} are used. We additionally use calibration plots and \textit{accuracy-rejection} plots to get a complete understanding of the UQ quality of these models.

ECE measures the absolute difference between the accuracy and the confidence of a prediction. It indicates how much the confidence of the model corresponds to its performance. Lower ECE values indicate a better calibration, where 0 is a perfect score.
Since accuracy can only be determined with multiple samples, samples are binned by their confidence in 10 bins. 
ECE is then calculated as
\begin{equation}
    ECE = \sum_{m=1}^{M} \frac{|B_m|}{N} \left| \text{acc}(B_m) - \text{conf}(B_m) \right| 
    ,
\label{equation:ECE}
\end{equation}
where $M$ is the number of bins, \(|Bm|\) is the number of samples that fall in bin $m$, and $N$ is the total number of samples. $Acc(B_m)$ and $conf(B_m)$ are respectively the accuracy and mean confidence of the predictions in a bin. All bins can also be visualised as a \emph{calibration plot} \cite{guo2017calibration}, to show whether patterns are consistent.

ECE does not tell us whether the miscalibration is because of overconfidence or underconfidence (or a mixture of both). We use Net Calibration Error (NCE) to determine the direction of miscalibration. It is computed in the same way as ECE, but without taking the absolute value of the error. A negative NCE means overconfidence, while a positive NCE means underconfidence \cite{groot2024confidence}. %

The Brier score measures how well the model's predicted confidence matches the real probabilities, considering both accuracy and uncertainty. 
A score of 0 is perfect, while a score of 1 is the worst. The Brier score is defined as
\begin{equation}
    \text{Brier}(y, \hat{y}) = N^{-1} \sum_{i=0}^{N} (y_i - \hat{y}_i)^2
    ,
\label{equation:Brier}
\end{equation}
where \(y_i\) is the true probability and \(\hat{y}_i\) is the predicted probability. Brier Score measures both accuracy and uncertainty calibration. A model with high accuracy and relatively poor calibration will still get a good Brier Score. %

\subsubsection{Rejection ability}
Good uncertainty estimation should be able to distinguish between correct and incorrect predictions. This would allow a model with good UQ to reject samples that are difficult, and only give predictions for samples in which it is confident. To evaluate this we use rejection-accuracy plots. These are created by sorting the test samples by model uncertainty, and rejecting the most uncertain samples. The accuracy is then evaluated on the remaining samples. By rejecting an increasing number of samples, a model with good UQ should be able to increase its accuracy. 

\section{Results}\label{sec:results_new} %

\begin{figure}[t]
    \begin{minipage}[t]{0.45\linewidth}
      \centering
      \centerline{\includegraphics[width=\linewidth]{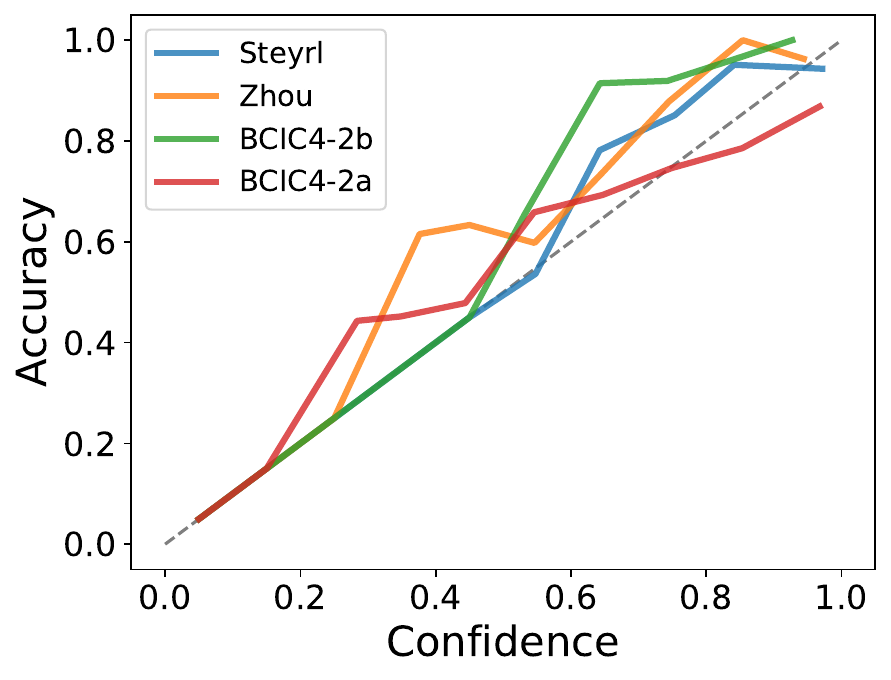}}
      \centerline{(a) MDRM}
    \end{minipage}\hfill
    \begin{minipage}[t]{0.45\linewidth}
      \centering
      \centerline{\includegraphics[width=\linewidth]{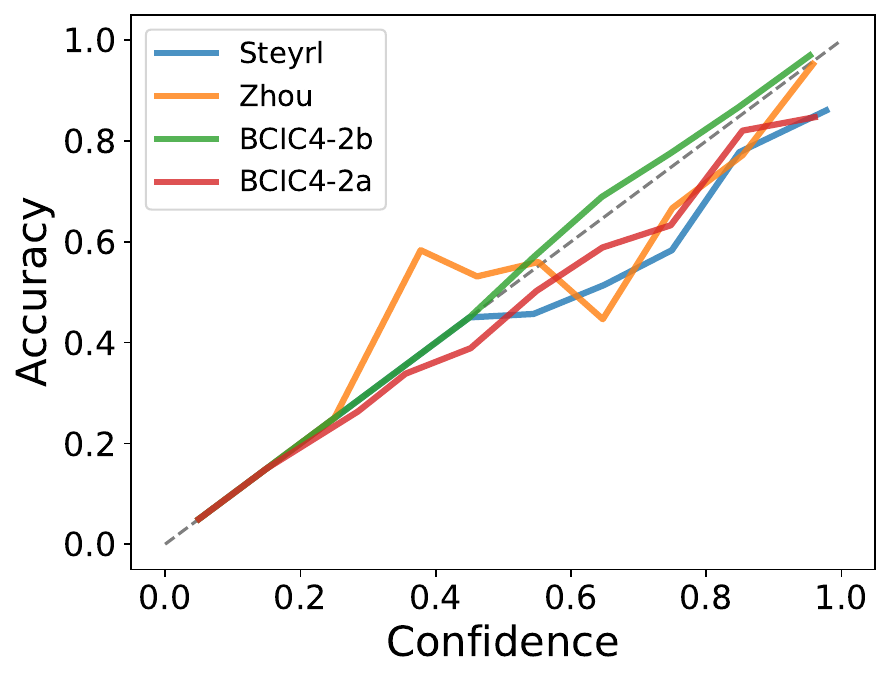}}
      \centerline{(b) MDRM-T}
    \end{minipage}
    
    \begin{minipage}[t]{0.45\linewidth}
      \centering
      \centerline{\includegraphics[width=\linewidth]{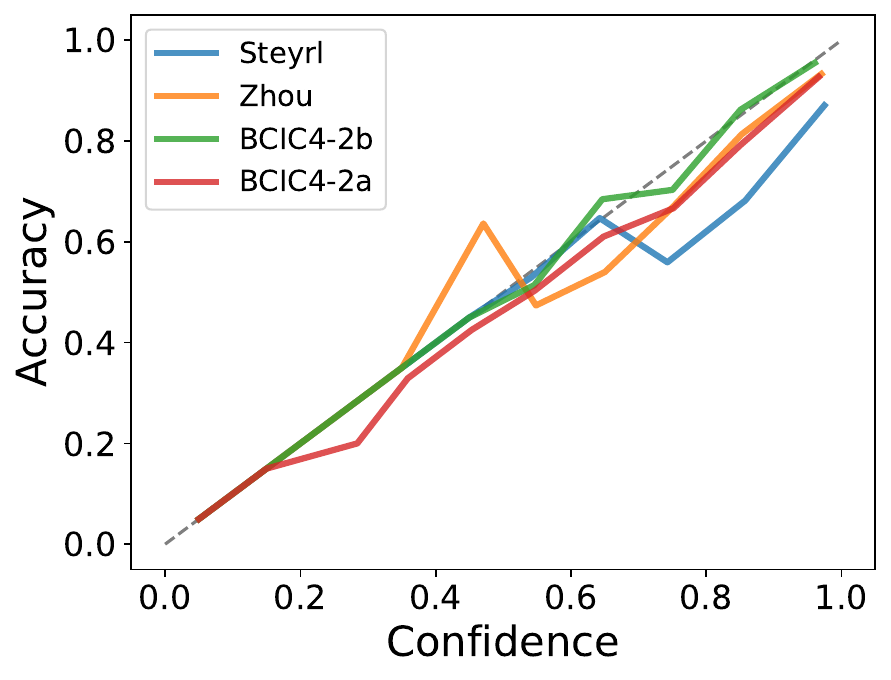}}
      \centerline{(c) CSP - LDA}
    \end{minipage}\hfill
    \begin{minipage}[t]{0.45\linewidth}
      \centering
      \centerline{\includegraphics[width=\linewidth]{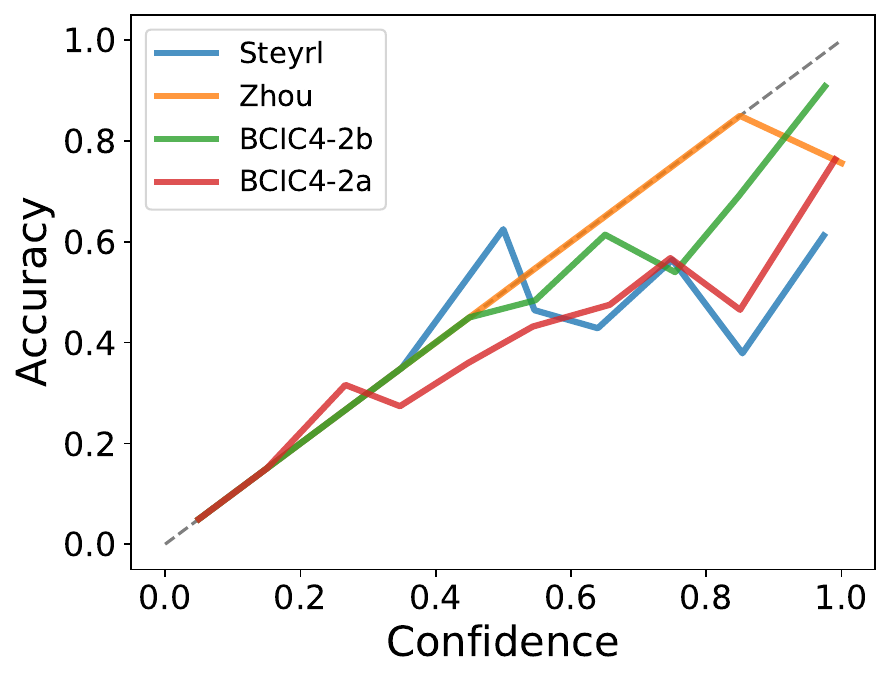}}
      \centerline{(d) DUQ}
    \end{minipage}
    
    \begin{minipage}[t]{0.45\linewidth}
      \centering
      \centerline{\includegraphics[width=\linewidth]{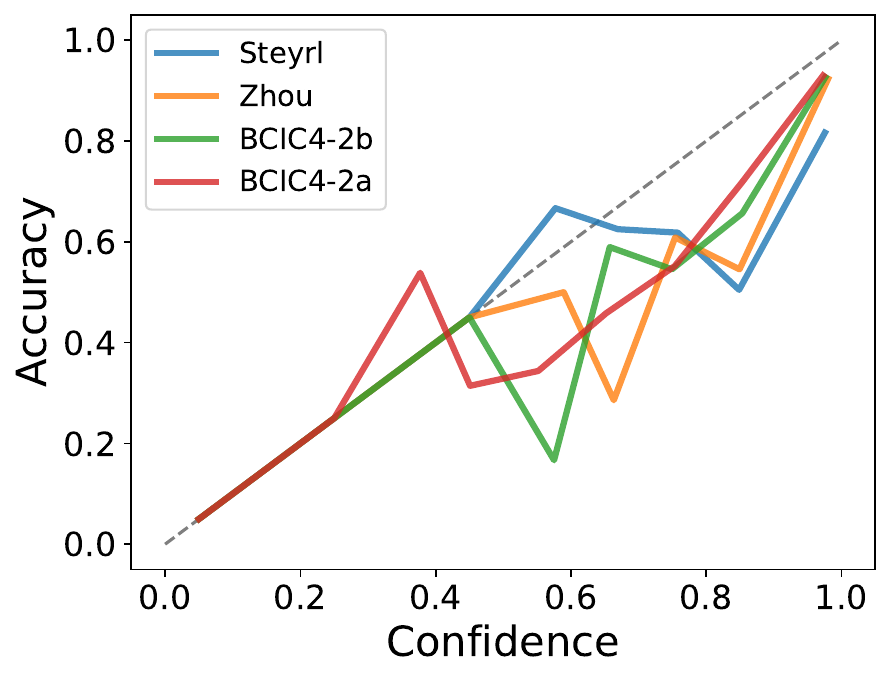}}
      \centerline{(e) Deep Ensembles}
    \end{minipage}\hfill
    \begin{minipage}[t]{0.45\linewidth}
      \centering
      \centerline{\includegraphics[width=\linewidth]{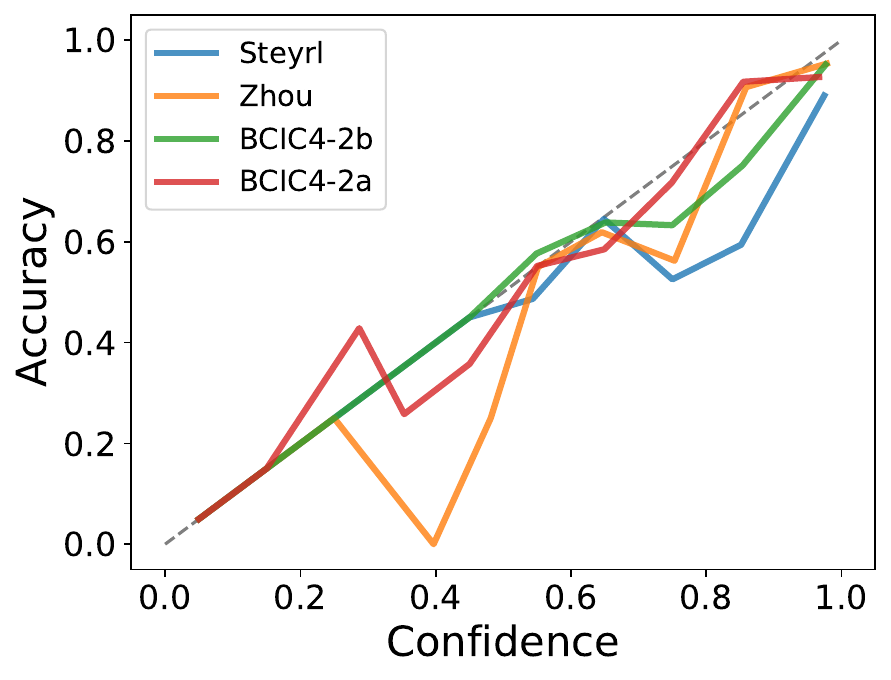}}
      \centerline{(f) CNN}
    \end{minipage}
    
    \caption{Calibration plots for the different models,  for each dataset. The diagonal line represents perfect calibration. We see that the Deep Learning methods (d-f) are all overconfident, while MDRM (a) is underconfident. MDRM-T (b) and CSP-LDA (c) give generally well-calibrated uncertainties.}\label{fig:calibration_plots}
\end{figure}

\begin{table*}[t]
\centering
\caption{Comparison of within-subject mean performance, with the standard deviation across subjects. \textbf{Boldface} indicates the best-performing model for a metric and dataset. Deep Learning methods (DE, CNN) give better classifications (Acc. \& Brier), but MDRM-T and CSP-LDA give better uncertainties (NCE \& ECE). Deep Learning is more computationally expensive.}

\resizebox{\linewidth}{!}{
\newlength{\charfill}
\setlength{\charfill}{0.55em}
\begin{tabular}{llrrrr r r}
\toprule
Metric & Dataset & MDRM & MDRM-T & CSP-LDA & DUQ & DE & CNN\\
\midrule
\multirow{4}{*}{\textbf{Acc. \%$\uparrow$}} & Steyrl      & 70.3 $\pm$ 17.0\%     & 70.3 $\pm$ 17.0\% &   \textbf{75.9 $\pm$ 15.5\%}  &  51.3 $\pm$ \hspace{\charfill}7.4\%   & 71.0 $\pm$ 14.7\%  & 70.8 $\pm$ 14.6\%             \\
                          & Zhou      & 72.3 $\pm$ \hspace{\charfill}7.1\%     & 72.3 $\pm$ \hspace{\charfill}7.1\%  & 77.6 $\pm$ \hspace{\charfill}8.5\%     &  76.7 $\pm$ \hspace{\charfill}2.3\%  &  83.0 $\pm$ \hspace{\charfill}3.5\%  & \textbf{84.3 $\pm$ \hspace{\charfill}5.7\%}            \\
                          & BCIC4-2b      & 71.4 $\pm$ 11.2\%     & 71.4 $\pm$ 11.2\% & 72.7 $\pm$ 11.2\%   &  77.1 $\pm$ 13.1\%   &  \textbf{79.1 $\pm$ 12.3\%}  & 79.0 $\pm$ 13.4\%            \\
                          & BCIC4-2a      & 58.2 $\pm$ 12.7\%     & 58.2 $\pm$ 12.7\%  & 66.5 $\pm$ 15.0\%     &  56.8 $\pm$ 19.8\%   &  \textbf{72.3 $\pm$ 18.2\%} & 71.8 $\pm$ 17.0\% \\\midrule
\multirow{4}{*}{\textbf{ECE $\downarrow$}} & Steyrl   & 0.163 $\pm$ 0.060   & \textbf{0.155 $\pm$ 0.087} & 0.231 $\pm$ 0.116   &  0.257 $\pm$ 0.132   & 0.276 $\pm$ 0.115   & 0.214 $\pm$ 0.064            \\
                          & Zhou      & 0.163 $\pm$ 0.033      & 0.148 $\pm$ 0.057 & \textbf{0.122 $\pm$ 0.042}      &  0.233 $\pm$ 0.026   &  0.264 $\pm$ 0.028 & 0.202 $\pm$ 0.042             \\
                          & BCIC4-2b      & 0.186 $\pm$ 0.102     & \textbf{0.066 $\pm$ 0.028} & 0.074 $\pm$ 0.039     &  0.164 $\pm$ 0.086   &  0.209 $\pm$ 0.081 & 0.121 $\pm$ 0.081             \\
                          & BCIC4-2a      & 0.156 $\pm$ 0.067     & 0.146  $\pm$  0.058 & \textbf{0.136 $\pm$ 0.040}   &  0.265 $\pm$ 0.062   &  0.198 $\pm$ 0.061 & 0.164 $\pm$ 0.061 \\\midrule

\multirow{4}{*}{\textbf{NCE $\rightarrow 0$}} & Steyrl      & 0.072 $\pm$ 0.102     & \textbf{-0.067 $\pm$ 0.129} & -0.112 $\pm$ 0.155       &  -0.206 $\pm$ 0.162   & -0.152 $\pm$ 0.125 & -0.102 $\pm$ 0.138      \\
                          & Zhou      & 0.122 $\pm$ 0.069     & \textbf{-0.024 $\pm$ 0.095}   & -0.057 $\pm$ 0.078    &  -0.233 $\pm$ 0.026   &  -0.210 $\pm$ 0.110   & -0.058 $\pm$ 0.126\\
                          & BCIC4-2b      & 0.186 $\pm$ 0.102     & 0.003 $\pm$ 0.003 & \textbf{-0.002 $\pm$ 0.048}     &  -0.069 $\pm$ 0.079   &  -0.189 $\pm$ 0.106     & -0.069 $\pm$ 0.109         \\
                          & BCIC4-2a      & 0.044 $\pm$ 0.124   & -0.045 $\pm$ 0.052 & -0.029 $\pm$ 0.075      &  -0.235 $\pm$ 0.097   &  -0.115 $\pm$ 0.076 & \textbf{-0.015 $\pm$ 0.117} \\\midrule
                          
                \multirow{4}{*}{\textbf{Brier $\downarrow$}} & Steyrl      & 0.183 $\pm$ 0.074     & 0.187  $\pm$ 0.093 & \textbf{0.176 $\pm$ 0.108}    &  0.311 $\pm$ 0.087   & 0.218 $\pm$ 0.097   & 0.204 $\pm$ 0.090            \\
                          & Zhou      & 0.142 $\pm$ 0.031    & 0.137 $\pm$ 0.035 & 0.104 $\pm$ 0.032     &  0.155 $\pm$ 0.017    &  \textbf{0.090 $\pm$ 0.021} & 0.079 $\pm$ 0.023             \\
                          & BCIC4-2b      & 0.213 $\pm$ 0.036     & 0.180 $\pm$ 0.050  & 0.172 $\pm$ 0.053     &  0.163 $\pm$ 0.072  &  0.148 $\pm$ 0.077 & \textbf{0.143 $\pm$ 0.083}            \\
                          & BCIC4-2a      & 0.141 $\pm$ 0.029    & 0.137 $\pm$ 0.034 & 0.110 $\pm$ 0.039      &  0.157  $\pm$ 0.044  &  0.103 $\pm$ 0.051 &  \textbf{0.095 $\pm$ 0.051}  \\
                          \midrule

\multicolumn{2}{l}{\textbf{Avg. Train time (s)}} & 0.08 & 0.08 & 0.78 & 32.34 & 141.82 & 28.37 \\
\multicolumn{2}{l}{\textbf{Avg. Inference time (ms)}}&  0.078 & 0.076 & 0.143 & 1.281 & 6.198 & 1.240 \\

                          \bottomrule
                          
\end{tabular}}
\label{table:model performances}
\end{table*}

We first observe the uncertainty calibration for the different models in Figure \ref{fig:calibration_plots}. We see that MDRM is generally underconfident (accuracy higher than confidence), while the Deep Learning methods are all overconfident. CSP-LDA is generally well-calibrated, and MDRM-T (with temperature scaling) is also well-calibrated. %

Table \ref{table:model performances} provides a comparison of all metrics of all models over all datasets. We see that the Deep Ensemble (DE) and the CNN generally give the highest accuracy with negligible difference between them. Only on the dataset from Steyrl \cite{steyrl2016random} the CSP-LDA gives better accuracy. We see that MDRM and MDRM-T have the same accuracies, because the classifications are not affected by Temperature Scaling.

Based on the ECE we can see MDRM-T and CSP-LDA give better calibrated uncertainty estimates for most datasets. For all datasets one of these two methods has the lowest ECE. They are comparatively well calibrated and (based on the NCE) we see that they are only minimally overconfident. These results show that MDRM-T solves the problem of underconfidence found in MDRM, resulting in only a slight overconfidence. This slight overconfidence might be because Temperature Scaling is optimised on the train data instead of a validation split. This leaves more data available for training, but could have resulted in a small amount of overconfidence. 

The Deep Learning based methods generally show worse calibration. DUQ and Deep Ensembles (DE) show large overconfidence on most datasets. The CNN is less overconfident, which is surprising considering that Deep Ensembles are typically considered to have better uncertainty estimation than standard CNNs \cite{manivannan2024uncertainty}.

In the Brier Score we see the combination of accuracy and uncertainty evaluated for the different models. We see that the models that get good accuracies also get good Brier Scores. We can see that MDRM-T gets better (lower) Brier Scores than MDRM, because the accuracy is the same but the uncertainty is better. We also see that the simple CNN sometimes gets better Brier Scores than the Deep Ensembles, because the Deep Ensembles only have slightly better accuracy, but much worse uncertainty estimation. 

We also consider the difference in computational time between the models, because Deep Learning models are sometimes considered a poor fit for online continual BCI systems because of their increased computational cost. We observed the average training time (per subject) and average inference time (per sample) using a MacBook M1 Pro (2021, 10-core CPU/16-core GPU, 32GB RAM). Deep Learning models can take 30-150 seconds to train while the non-Deep Learning models are trained within one second. Inference time per sample for Deep Learning is over 1 millisecond, while the non-Deep Learning models take less than 0.1 millisecond. This shows that Deep Learning systems are much more computationally expensive, but on a modern high-end laptop using ShallowConvNet \cite{schirrmeister2017deep} the extra training time is not prohibitive, and the extra inference time is not a problem for standard setups.

\begin{figure}[t]
    \begin{minipage}[t]{0.45\linewidth}
      \centering
      \centerline{\includegraphics[width=\linewidth]{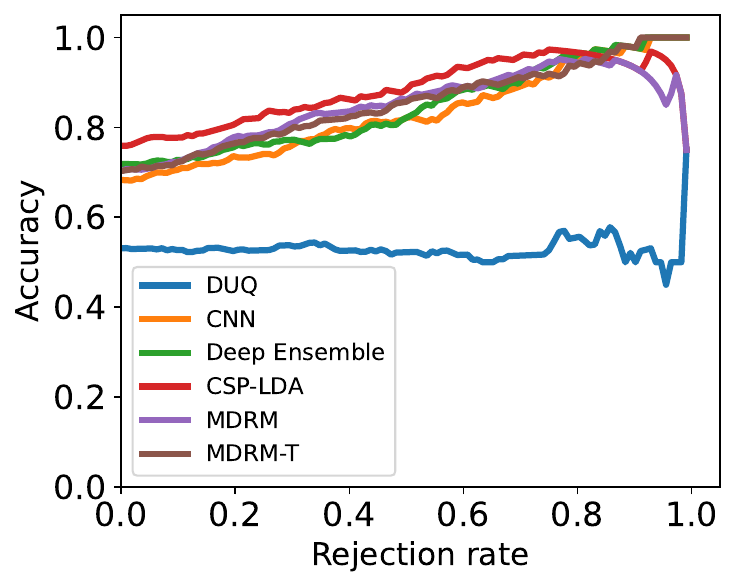}}
      \centerline{(a) Steyrl}
    \end{minipage}\hfill
    \begin{minipage}[t]{0.45\linewidth}
      \centering
      \centerline{\includegraphics[width=\linewidth]{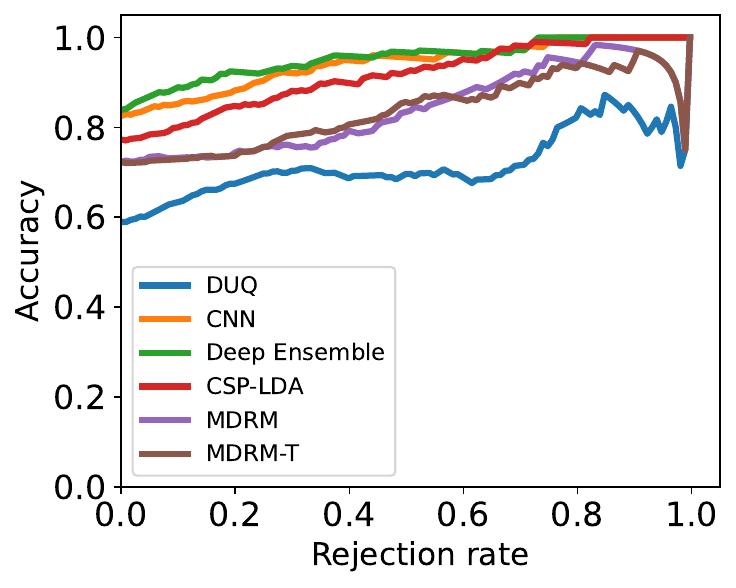}}
      \centerline{(b) Zhou}
    \end{minipage}
    
    \begin{minipage}[t]{0.45\linewidth}
      \centering
      \centerline{\includegraphics[width=\linewidth]{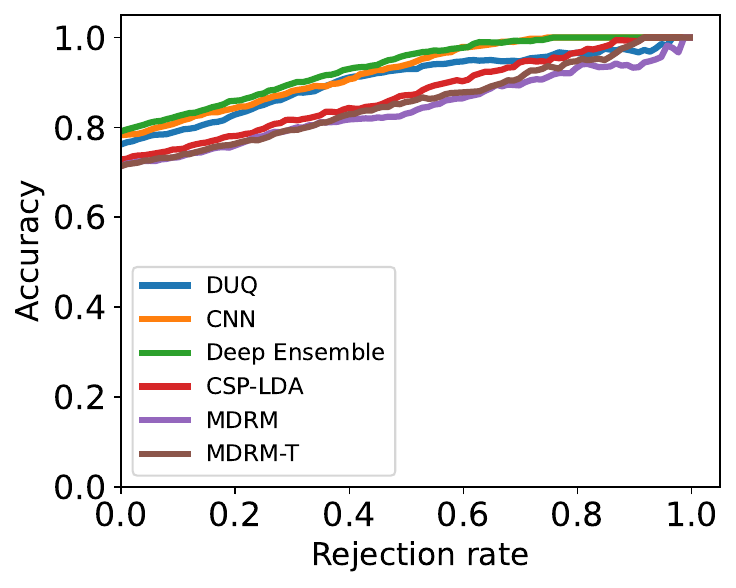}}
      \centerline{(c) BCIC4-2b}
    \end{minipage}\hfill
    \begin{minipage}[t]{0.45\linewidth}
      \centering
      \centerline{\includegraphics[width=\linewidth]{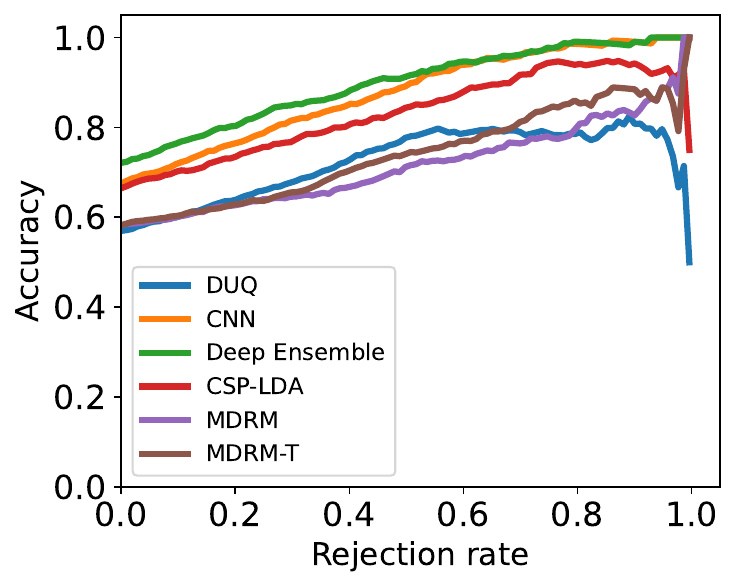}}
      \centerline{(d) BCIC4-2a}
    \end{minipage}
    
    \caption{Rejection rate (percentage of samples below an uncertainty threshold) and accuracy (calculated on remaining samples) for each model on the different datasets. By rejecting the most uncertain samples all models can improve their accuracy. The CNN achieves the best combination of high accuracies with low rejection rates.}\label{fig:rejection-accuracy}
    \label{fig:res}
\end{figure}

Lastly we consider the ability of the uncertainty to distinguish between easy and difficult samples. This is done through the rejection-accuracy plots shown in Figure \ref{fig:rejection-accuracy}. We can see that by rejecting the most uncertain samples all models are able to achieve higher accuracies on all datasets. %

In the BCIC4-2a dataset we can see that MDRM-T goes from an accuracy of less than 60\% to over 80\% by rejecting 70\% of the samples. This can make a BCI that might otherwise be experienced as inaccurate give much more reliable predictions. The model then does not always make a prediction, but when a prediction is made it is likely to be correct. 

We see that the quality of the final model is largely a function of the original accuracy. %

\section{Discussion}\label{sec:discussion}

We have compared the Uncertainty Quantification capabilities of non-Deep Learning methods that are common for Motor Imagery BCIs against Deep Learning methods especially designed to improve uncertainty estimation. We see variations over different datasets, but we generally find that while the Deep Learning methods showed better accuracies, the BCI Machine Learning methods showed better uncertainty estimation. CSP-LDA has good uncertainty estimation. We found that MDRM is \textit{underconfident}, but our introduction of MDRM-T with Temperature Scaling solves this underconfidence. 

The Deep Learning methods all present as overconfident, which is in line with what has been shown in other domains. When considering a practical application of uncertainty we show that by being selective on which samples should get a prediction all models are able to achieve better accuracies. Being selective about inference is shown to improve accuracy, but also makes it so the model does not always give a prediction. Future work can implement this rejection system in an online study, where we can observe whether a rejection system can improve the learning effects during Motor Imagery by giving more reliable feedback. A selective but precise BCI may also give a better user experience as the user would feel more in control of the resulting commands. Our results show that for Deep Learning models, such reject systems cannot trust the true probability, but for CSP-LDA or MDRM-T those probabilities are fairly reliable.

\subsection{Limitations}

While this study provides an analysis of the performances of four different UQ methods, several limitations should be considered when adapting this study's findings. 

The study does not run experiments with data that is out of distribution \cite{chetkin2023bayesian}. This means that the ML models tested in this study are not scored on how well they perform on UQ performance when data is suddenly very different from what it was trained on. In real-life scenarios, however, this can happen when a BCI receives artifacts or off-task EEG data. It would be interesting for future works to evaluate the performances of the models in detecting out-of-distribution data and see if the same models still perform best. 

Additionally, further analysis and experimentation is needed to understand how rejection systems affect the user experience. This requires further investigation of the distribution and properties of rejected samples and how rejection affects the subjective experience of control. 

The scope of this study is limited to Motor-Imagery BCIs. It may be realistic to assume that these findings could extend to other BCI paradigms, but this cannot be concluded from our findings.

\section{Conclusion}\label{sec:conclusion}
Our paper introduced MDRM-T, where MDRM is combined with temperature scaling to get better calibrated uncertainty estimates. We show that it gives better calibrated uncertainty estimates than MDRM, without affecting the classification performance. We also found that standard BCI models give better uncertainty estimates than Deep Learning models and are much more computationally efficient, but Deep Learning gave better classification performances. 
 
This study demonstrates how to systematically analyze in-distribution uncertainty estimation performance in Motor-Imagery BCIs, looking at ECE, NCE, Brier Score and rejection ability. This is a clear and simple setup that can be applied to future models to ensure the predicted class probabilities are reliable. This setup is task-agnostic and evaluates uncertainty in general, so the findings from the experiments that we show here can form a basis for specific uncertainty quantification tasks for Motor Imagery BCIs \cite{manivannan2024uncertainty, chetkin2023bayesian}.

\bibliographystyle{IEEEbib}
\bibliography{literature_bachelorthesis, strings, refs}

\end{document}